\documentclass[11pt,letterpaper]{article}
\usepackage{naaclhlt2016}
\usepackage{times}
\usepackage{latexsym}

\usepackage{helvet}
\usepackage{courier}
\usepackage{graphicx}
\usepackage{amsmath,amsfonts,amsthm}
\usepackage{epstopdf}
\usepackage{url}
\naaclfinalcopy 
\def\naaclpaperid{***} 

\title{Category Enhanced Word Embedding}

\author{{Chunting Zhou}$^{1}$, Chonglin Sun$^{2}$, Zhiyuan Liu$^{3}$, Francis C.M. Lau$^{1}$\\
Department of Computer Science, The University of Hong Kong$^{1}$\\
School of Innovation Experiment, Dalian University of Technology$^{2}$\\
Department of Computer Science and Technology, Tsinghua University, Beijing$^{3}$\\
}

\begin{document}

\maketitle

\begin{abstract}
Distributed word representations have been demonstrated to be effective
in capturing semantic and syntactic regularities. Unsupervised
representation learning from large 
unlabeled corpora can 
learn similar representations for those words that present similar
co-occurrence statistics. Besides local occurrence statistics, global
topical information is also important knowledge that may help
discriminate a word from another. In this paper, we incorporate
category information of documents in the learning of word
representations and to learn the proposed models in a document-wise
manner. Our models outperform several
state-of-the-art models in word analogy and word similarity tasks.
Moreover, we evaluate the learned word vectors on sentiment analysis
and text classification tasks, which shows the superiority of our
learned word vectors.
We also learn high-quality category embeddings that reflect
topical meanings.
\end{abstract}

\section{Introduction}
Representing each word as a dense real-valued vector, also known as
word embedding, has been exploited extensively in NLP communities
recently~\cite{Bengio:03,C&W:08,M&H:2008,Socher:2011,Mikolov:13a,glove:14}.
Besides addressing the issue of dimensionality, word embedding also
has the good property of generalization. Training word vectors from
a large amount of data 
helps learn the intrinsic statistics of languages.
A popular approach to training a statistical language model
is to build a simple neural network architecture with an
objective to maximize the probability of predicting a word given
its context words. After the training has converged, words with similar meanings are
projected into similar vector representations and linear regularities
are preserved.\\
\indent Distributed word representation learning based on local
context windows could only capture semantic and syntactic similarities
through word neighborhoods. Recently, instead of purely unsupervised
learning from large 
corpora, 
linguistic knowledge such as semantic and syntactic knowledge have
been added to the training process. Such additional knowledge could define
a new basis for word representation, enrich input information, and serve
as complementary supervision when training the neural
network~\cite{Bian:14}. For example, Yu and Dredze~\shortcite{Yu:14}
incorporate relational knowledge in their neural network model to
improve lexical semantic embeddings.\\
\indent Topical information is another kind of knowledge that appears
to be also attractive for training more effective word embeddings. Liu et al~\shortcite{Liu:15} leverage implicit
topics generated by LDA to train topical word embeddings for
multi-prototype vectors of each word.
Co-occurrence of words within local context windows provides partial
and basic statistical information between words; however, words in
different documents with dissimilar topics may show different 
categorical properties. For example, ``cat" and ``tiger" are likely to
occur under the same category of ``Felidae" (from Wikipedia) but less likely to occur
within the same context window. It is important for a word to know
the categories of its belonging documents when the neural network is
trained on large corpora. \\
\indent In this work, we propose to incorporate explicit document
category knowledge as additional input information and also as auxiliary
supervision. 
WikiData is a document-based corpus 
where each document is
labeled with several categories.
We leverage this corpus to train
both word embeddings and category embeddings in a document-wise
manner. Generally, we represent each category as a dense
real-valued vector which has the same dimension as word
embeddings in the model. We propose two models for integrating
category knowledge, namely {\bf category enhanced word embedding}
(CeWE) and {\bf globally supervised category enhanced word
embedding} (GCeWE).
In the well-known
CBOW~\cite{Mikolov:13a} architecture, each middle word is predicted by
a context window, which is convenient for plugging category
information into the context window when making predictions. In the
CeWE model, we find that with local additional category
knowledge, word embeddings 
outperform CBOW and
GloVe~\cite{glove:14} 
significantly in word similarity tasks. In the GCeWE model, based on
the above local reinforcement, we
investigate predicting corresponding categories using words in a
document after the document has been trained through a local-window model.
Such auxiliary supervision can be viewed as a global constraint at the
document level. We also demonstrate that by combining additional local
information and global supervision, the learned word embeddings 
outperform CBOW and GloVe 
in the word analogy task~\cite{Mikolov:13a}.\\
\indent Our main contribution is that we 
integrate explicit category
information into the learning of word representation to train
high-quality word embeddings. 
The resulting category
embeddings also capture the semantic meanings of topics.
\begin{figure}[!hbt]
\centering
\includegraphics[width=2in]{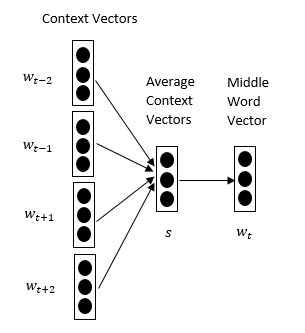}
\caption{The CBOW architecture that predicts the middle word using the average context window vectors.}
\end{figure}
\section{Related Work}
Word representation is a key component of many NLP and IR related
tasks. The conventional representation for words known as
``bag-of-words" (BOW) ignores the word order and suffers from
high dimensionality, and reflects little relatedness and distance
between words. Continuous word embedding was first proposed in
~\cite{1986} and has become a successful representation method in
many NLP applications including machine translation~\cite{zou},
parsing~\cite{Socher:2011}, named entity
recognition~\cite{passos}, sentiment analysis~\cite{Glorot},
part-of-speech tagging~\cite{Collobert:11} and text
classification~\cite{Le:14}.\\
\indent Many prior works have explored how to learn effective word
embeddings that can capture the words' intrinsic similarities and
discriminations. Bengio et al.~\shortcite{Bengio:03} proposed to train
an n-gram model using a neural network architecture with one hidden
layer, and obtained good generalization. In ~\cite{M&H:07}, Minh and
Hinton proposed three new probabilistic models in which they used binary
hidden variables to control the connection between 
preceding words and the next word.\\
\indent The methods mentioned above require high computational cost. To
reduce the computational complexity, softmax models with hierarchical
decomposition of probabilities~\cite{M&H:2008,hier} have been proposed to
speed up the training and recognition. More recently, Mikolov et al.
~\shortcite{Mikolov:13a,Mikolov:13b} proposed two models---CBOW and
Skip-Gram---with highly efficient training methods to learn
high-quality word representations; they adopted a negative sampling
approach as an alternative to the hierarchical softmax. Another
example that explored the
co-occurrence statistics between words is GloVe~\cite{glove:14},
which combines
global matrix factorization and local context window methods.\\
\indent The above models exploit word correlations within context
windows; however, several recently proposed models explored how to
integrate other sources
of knowledge into word representation learning. For example, Qiu et
al.~\shortcite{siyu} incorporated morphological knowledge to help learn
embeddings for rare and unknown words.\\
\indent In this work, we design models to incorporate document
category information into the learning of word embeddings where the
objective is to correctly predict a word with not only context words
but also its category knowledge. We show that word embeddings
learned with document category
knowledge have better performance in word similarity tasks and
word analogical reasoning tasks. Besides, we also evaluate the
learned word embeddings on text classification tasks and show the
superiority of our models.
\begin{figure}[!hbt]
\centering
\includegraphics[width=2.5in]{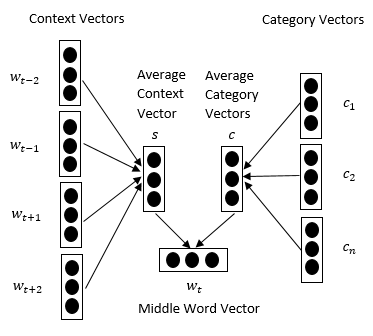}
\caption{Category enhanced word embedding architecture that predicts the middle word using both context vectors and category vectors.}
\end{figure}

\section{Methods}
In this section, we show two methods of integrating document
category knowledge into the learning of word embeddings. First, we
introduce the CeWE model where the
context vector for predicting the middle word is enriched with
document categories. Next, based on CeWE, we introduce the GCeWE model where
word embeddings and category embeddings are jointly trained under a
document-wise global supervision on words within a document.

\subsection{Category Enhanced Word Embedding}
In this section, we present our method for training word embeddings
and category embeddings jointly within local windows. We extend the
CBOW~\cite{Mikolov:13a} architecture by incorporating 
category information of each document, to learn more comprehensive and
enhanced word representations. The architecture of the CBOW model is
shown in Figure 1 and its objective is to maximize the log probability
of the current word $t$, given its context window $s$:
\begin{align}
\emph{J}(\theta) =\sum_{t=1}^{V}{\sum_{s\in context(t)}{\log(p(t|s))}}
\end{align}
where $V$ is the size of the word vocabulary and $context(t)$ is the
set of observed context windows for the word $t$. CBOW basically
defines the probability $p(t|s)$ using the softmax function:
\begin{align}
p(t|s) = \frac{\exp(w^{'T}_tv_s)}{\sum_{j \in V, j \neq t}{\exp(w^{'T}_jv_s)}}
\end{align}

where $w^{'}_t$ is the output word vector of word $t$. Meanwhile, each
word $t$ is maintained with an input word vector $w_t$. And a context
window vector $v_s$ is usually formulated as the average of context
word vectors $\frac{1}{2k}\sum_{{t-k} \leq j \leq {t+k},j\neq t}^{
}{w_j}$, where 
$k$ is the size of the window to the left and
to the right. Mikolov~\shortcite{Mikolov:13a,Mikolov:13b} also
proposed some efficient techniques including hierarchical softmax and
negative sampling to replace full softmax during the optimization
process.\\
\indent Context window based models are prone to suffer from the
lack of global information. Except for those frequently used words
such as function words ``he", ``what", etc., most words are used commonly
under some certain language environment. For example, ``rightwing"
and ``anticommunist" occur most likely under politically related
topics; the football club ``Millwall" occurs most likely under
football related topics. To make semantically similar words behave
more closely within the vector space, we propose to take advantage of
the topic background in which the words lie during training. Different
from the CBOW model, we plug in the category information to align word
vectors under the same topic more closely and linearly when predicting the
middle word, which is as shown in Figure 2. To train this model, we
create a continuous real-valued vector for each category. The
dimension of the category vector is set to be the same as the word vector.
Since the number of categories for each document is not fixed, we
denote the last category vector in Figure 2 as $c_n$. We train the
CeWE model in a document-wise manner instead of
taking the entire corpus as a sequence of words. In this way, we
utilize the Wikipedia dumps which have associated each document with
multiple categories. The creation of our dataset is described in
details in Section 4.1.\\
\indent We combine the average of the context window vector $v_s$ together
with the weighted average of the category vectors to act as the new
context vectors. Let $c_i$ denote the vector for the $i^{th}$
category, and $category(m)$ the set of categories for the $m^{th}$
document. The new objective function is then:
\begin{align}
\emph{J}(\theta) =\sum_{t=1}^{V}{\sum_{s\in context(t)}{\log(p(t|s,u))}}
\end{align}
where the current context window $s$ belongs to document $m$. The
probability $p(t|s,u)$ of observing the current word $t$ given its
context window $s$ and document categories $u$ is defined as follows:
\begin{align}
p(t|s, u) = \frac{\exp(w_t^{'T}(v_s + \lambda z_u))}{\sum_{j \in V, j \neq t}{\exp(w_{j}^{'T}(v_s + \lambda z_u))}}
\end{align}
where $z_u$ is the document category representation formulated as the
average of category vectors $\frac{1}{|category(m)|}\sum_{i \in
category(m)}{c_i}$, and $\lambda$ is a hyperparameter to control the weight
of the category vectors which play a role in predicting the middle word. We
make use of negative sampling to optimize the objective function (3).

\subsection{Globally Supervised CeWE}
In the above model, we only integrate category information
into local windows, enforcing inferred words to capture topical
information and pulling word vectors under the same topic closer.
However, an underlying assumption that can be easily seen is that the
distribution of document representations should be in accordance with the
distribution of categories. Thus, based on CeWE, we use the document
representation to predict the corresponding categories as a global
supervision on words, resulting in our GCeWE model.

\subsubsection{Model Description}
The objective of GCeWE has two parts: the first one is the same as
that of the CeWE model,
and the other one is to maximize the log probability of observing document
category $i$ given a document $m$, as follows:
\begin{align}
\emph{J}(\theta) = \nonumber \sum_{t=1}^{V}{\sum_{s\in context(t)}{\log(p(t|s,u))}} ~+ \\
 \sum_{m=1}^{M}{\sum_{i\in category(m)}{\log(p(i|m))}}
\end{align}
Similarly, $p(i|m)$ is defined as:
\begin{align}
p(i|m) = \frac{\exp(c_{i}^Td_m)}{\sum_{j \in C, j \neq i}{\exp(c_{j}^Td_m)}}
\end{align}
where $C$ is the size of all categories, $d_m$ denotes the document
representation of the $m^{th}$ document and $i \in category(m)$.\\
\indent Another problem to be solved is how to effectively represent a
document to make the document representation discriminative. 
From experiments we find that with either average or TF-IDF weighted
document representation that involves all words in a document, word embeddings trained by the GCeWE
model shows little superiority in the word analogy task.
We conjecture that the average operation makes the document
representation less discriminative so that the negative
sampling method could not sample informative negative categories, as
we discuss below.\\
\indent It has been shown that the TF-IDF value is a good measure of
whether a word is closely related to the document topics. Therefore,
before imposing the global supervision on the document representation,
we first calculate the average TF-IDF value of all words in a document denoted
as $\text{AVGT}$, and we select words that have a TF-IDF value larger than $\text{AVGT}$
to participate in the global supervision. Instead of an average
operation on these selected words, we use each of these words to
predict the document categories separately. Thus, our new objective
function becomes:
\begin{align}
\emph{J}(\theta) = \nonumber \sum_{t=1}^{V}{\sum_{s\in context(t)}{\log(p(t|s,u))}} ~+ \\
 \sum_{m=1}^{M}\sum_{l \in L_m}{{\sum_{i\in category(m)}{\log(p(i|l))}}}
\end{align}
where $L_m$ is the set of words selected from the $m^{th}$ document
according to $\text{AVGT}$. The probability of observing a category $i$ given
a selected word $l$ is defined similarly to Equation (6), as below:
\begin{align}
p(i|l) = \frac{\exp(c_{i}^Tw_l)}{\sum_{j \in C, j \neq i}{\exp(c_{j}^Tw_l)}}
\end{align}

\subsubsection{Optimization with Adaptive Negative Sampler}
We also adopt the efficient negative sampling as in
~\cite{Mikolov:13b} to maximize the second part of the objective
function. For positive samples, we 
rely on the document representation to
predict all categories of its belonging document. To select the most
``relevant" negative category samples that could help accelerate the
convergence, we employ the adaptive and context-dependent negative
sampling proposed in ~\cite{Steffen:14} for pairwise learning. Steffen
and Freudenthaler's sampling method aims to sample the most
informative negative samples for a given user and it works well in
learning recommender systems where the target is to recommend the most
relevant items for a user. It is analogous to selecting the most
informative negative categories for a document. Note that the category
popularity has a tailed distribution: only a small subset of
categories have a high occurring frequency while the majority of
categories do not occur very often at all. SGD algorithms with samples that
have a tailed distribution may suffer from noninformative negative
samples when using a uniform sampler. Noninformative samples have no
contribution to the SGD algorithm, as shown in ~\cite{Steffen:14},
which slow down the convergence.\\
\indent We employ the adaptive nonuniform sampler of
~\cite{Steffen:14} by regarding each word as a context and each
category as an item under the matrix factorization (MF) framework.
Elements of word vectors and category vectors can be viewed as a
sequence of factors. According to a sampled factor of the document
representation, we sample negative categories that should not
approximate the document representation in the vector space.\\
\indent We will show that with GCeWE the semantic word analogy
accuracy is improved remarkably as compared with the CBOW model.

\section{Experiments}
\subsection{Datasets }
WikiData is a document-oriented database, which is suitable for our
training methodology. We extract document contents and categories from
a 2014 Wikipedia dump.
Each document is associated with several categories.
As both the number of documents and that of categories are very large, we only
reserve documents with category tags corresponding to the
top $10^5$ most frequently
occurring categories. We note that there are many redundant
meaningless category entries like ``1880 births", ``1789 deaths", etc.,
which usually consist of thousands of documents from different fields
under one category. Although we cannot exclude all noisy categories,
we eliminate a fraction of these categories by some rules, resulting
in 86,664 categories and 2,271,411 documents. These categories occur
in the entire dataset 152 times on average. We also remove all stop
words in a predefined set from the
corpus. Besides, in our experiment, we remove all the words that occur
less than 20 times. Our final training data set has 0.87B
tokens and a vocabulary of 533,112 words.

\subsection{Experiment Settings and Training Details}
We employ stochastic gradient descent (SGD) for the optimization using
four threads on a 3.6GHz Intel i7-4790 machine. We randomly select
100,000 documents as held-out data for tuning hyperparameters and
use all documents for training.
The dimension of word vectors is chosen to be 300 for all
models in the experiment, and so the dimension of category vectors is
also 300. 20 negative words are sampled in the negative
sampling of CeWE and 20 negative categories
are sampled in the adaptive negative sampling of GCeWE. Different
learning rates are used when the category acts as additional input and
the supervised target and are denoted $\alpha$ and $\beta$ respectively.
We set $\alpha$ to be 0.02 and $\beta$ 0.015. We also use subsampling of
frequent words as proposed in ~\cite{Mikolov:13b} with the parameter of
1e-4. For the hyperparameter $\lambda$, we set it to be $1/cw$ where
$cw$ is the number of words within a context window. To make a fair
comparison, we train all models except GloVe for two epochs. In each
epoch, the dataset is gone through once in its entirety.\\
\indent The adaptive nonuniform negative sampling in the GCeWe model
involves two sampling steps: one is to sample an importance factor $f$
from all factors of a given word embedding and the other one is to sample
a rank $r$ from 300 factor dimensions. We draw a factor given a word
embedding from $p(f|w) \propto |w_f|\sigma_f$ where $w_f$ is the
$f^{th}$ factor of word vector $w$ and $\sigma_f$ is the standard
deviation of factor $f$ over all categories. A factor with a smaller
rank over all factors 
has greater weights than other factors. To
sample a smaller rank $r$, we draw $r$ from a geometric distribution
$p(r) \propto \exp(-r/\lambda)$ which has a tailed distribution. And in
our experiment, $\lambda = 5$.
\begin{table*}[t]
\begin{center}
\begin{tabular}{|l|r|r|r|r|r|r|r|}
\hline
\bf Model &\bf Corpus Size &\bf win\_size &\bf WS353 &\bf SCWS &\bf MC &\bf RG &\bf RW \\
\hline
Skip-gram-300d & 0.87B & 5 & 70.74 & 65.77 & 81.82 & 80.33 & 43.06  \\
\hline
Skip-gram-300d & 0.87B & 10 & 69.75 & 63.85 & 81.13 & 80.08 & 42.80 \\
\hline
CBOW-300d      & 0.87B & 10 & 67.62 & 65.77 & 81.00 & 81.17 & 41.10\\
\hline
CBOW-300d      & 0.87B & 12 & 68.41 & 65.72 & 81.86 & 82.20 & 41.46\\
\hline
CBOW-300d      & 0.87B & 14 & 68.99 & 65.52 & 81.57 & 82.82 & 41.47\\
\hline
CeWE-300d    & 0.87B & 10 & 72.78 & \bf 65.79 & 81.38 & 82.78 & 45.26\\
\hline
CeWE-300d    & 0.87B & 12 & 73.29 & 65.31 & 83.22 & \bf 84.51 & 45.90\\
\hline
CeWE-300d    & 0.87B & 14 & \bf 74.38 & 64.63 & \bf 84.58 & 83.61 & \bf 46.21\\
\hline
Glove-300d        & 0.87B & 10 & 68.28 & 59.22 & 75.30 & 77.30 & 37.00 \\
\hline
\end{tabular}
\end{center}
\caption{\label{sim-table} Spearman rank correlation $\rho \times 100$
on word similarity tasks. Scores in bold are the best ones in each
column.}
\end{table*}
\subsection{Evaluation Methods}
{\bf Word Similarity Tasks.} ~~The word similarity task is a basic
method for evaluating word vectors. We evaluate the CeWE model on five
datasets including WordSim-353~\cite{wordsim353}, MC~\cite{MC},
RG~\cite{RG}, RW~\cite{RW}, and SCWS~\cite{SCWS}, which contain 353,
30, 65, 2003, 1762 word pairs respectively. We use SCWS to evaluate
our word vectors without context information. In these datasets, each
word pair is given a human labeled correlation score according to the
similarity and relatedness of the word pair. We compute the spearman
rank correlation between the similarity scores calculated based on
word embeddings and human labeled scores.\\
{\bf Word Analogy Task.} ~~The word analogy task was first introduced
by Mikolov~\shortcite{Mikolov:13a}. It consists of analogical
questions in the form of ``a is to b as b is to \_\_?". The dataset
contains two categories of questions: 8869 semantic questions and
10675 syntactic questions. There are five types of relationships in
the semantic questions including capital and city, currency,
city-in-state, man and woman. For example, ``brother is to sister as
grandson is to \_\_?" is a question for ``man and woman". And there are
nine types of relationships in the syntactic questions including adjective
to adverb, opposite, comparative, etc. For example, ``easy is to
easiest is lucky is to \_\_?" is one question of ``superlative". We
answer such questions by finding the word whose word embedding $w_d$
has the maximum cosine distance to the vector ``$w_b - w_a + w_c$".\\
{\bf Sentiment Classification and Text Classification} We evaluate
the learned embeddings on two dataset: the IMDB ~\cite{maas2011} and
20NewsGroup \footnote{\url{http://qwone.com/~jason/20Newsgroups/.}}.
IMDB is a benchmark dataset for binary sentiment classification
which contains 25K highly polar movie reviews for training and 25K
movie reviews for testing. 20NewsGroup is a dataset of around 20000
documents organized into 20 different newsgroups. We use the
``bydate" version of 20NewsGroup, which splits the dataset into 11314
and 7532 documents for training and testing respectively. We choose
LDA, TWE-1 ~\cite{Liu:15}, Skip-Gram, CBOW, and GloVe as baseline
models. LDA represents each document as the inferred topic
distribution. For Skip-Gram, CBOW, GloVe and our models, we simply
represent each document by aggregating embeddings of words that have
a TF-IDF value larger the $\text{AVGT}$ and use them as document
features to train a linear classifier with Liblinear
\cite{liblinear}. For TWE-1, the document embedding is represented
by aggregating all topical word embeddings as described in
~\cite{Liu:15}, and the length of topical word embedding is double
that of word embedding or topic embedding. We set the dimension of both
word embedding and topic embedding in TWE-1 to be 300.

\begin{table*}[t]
\begin{center}
\begin{tabular}{|l|c|r|r|r|r|}
\hline
\bf Model & \bf win size & \bf Sem.(\%) & \bf Syn.(\%) & \bf Tot.(\%) \\
\hline
Skip-gram & 5 & 75.83 & 60.25 & 67.53  \\
\hline
Skip-gram & 10 & 76.25 & 58.21 & 66.64 \\
\hline
CBOW     &  10 & 74.87 & 62.44 & 68.15 \\
\hline
CBOW     & 12 & 75.08 & 62.43 & 68.34 \\
\hline
CBOW    & 14 & 73.90 & 62.34 & 67.74 \\
\hline
CeWE  & 10 & 72.71 & \bf 65.44 & 68.84 \\
\hline
CeWE  & 12 & 73.76 & 64.40 & 68.77 \\
\hline
CeWE  & 14 & 74.39 & 64.07 & 68.89 \\
\hline
GCeWE  & 10 &  \bf 76.56 & 65.14 & \bf 70.46\\
\hline
GloVe   & 10 & 75.36 & 63.42 & 69.00 \\
\hline
\end{tabular}
\end{center}
\caption{\label{sim-table} Results on word analogical reasoning task. }
\end{table*}
\subsection{Results and Analysis}
For word similarity and word analogical reasoning tasks, we compare our models with CBOW, Skip-Gram and the state-of-the-art
GloVe model. GloVe takes advantage of the global
co-occurrence statistics with weighted least square. All models
presented are trained using our dataset. For GloVe, we set the model
hyperparameters as reported in the original paper, which have achieved
the best performance. CBOW and Skip-Gram
are trained using the word2vec tool \footnote{\url{http://code.google.com/p/word2vec/}}. We first present
our results on word similarity tasks in Table 1 where the CeWE model
consistently achieves the best performance on all five datasets. This indicates that additional category information helps to learn high-quality word
embeddings that capture more precisely the semantic meanings. We also find
that as the window size increases, the CeWE model performs better for some
similarity tasks. 
The reason probably is that when the window size becomes
larger, more information of the context is added to the input vector,
and the additional category information enhances the contextual
meaning. However, the performance decreases as the window size exceeds 14.\\
\indent Table 2 presents the results of the word analogy task.
The CeWE model performs better than the CBOW model with additional
category information. By applying global supervision, the GCeWe
model outperforms CeWE and GloVe in this task.
We also observe that CeWE performs better in the word analogy task
when using larger window size, but GCeWE model has a better
performance when the window size is 10. So we only report the result
of GCeWE with window size of 10. Also, we note that GCeWE performs
worse compared to CeWE in word similarity tasks but better than CBOW
and the Skip-Gram model, and so we only report the result of the CeWE model
for the word similarity tasks.\\
\indent Table 3 presents the results of the tasks of sentiment
classification and text classification, and it is evident that
document representations
computed by our learned word embeddings consistently outperform
other baseline models. Although the documents are represented by
discarding word orders, they still show good performance in the document
classification tasks. This indicates that our models can learn
high-quality word embeddings with category knowledge. Moreover, we
can see that GCeWE performs better than CeWE on these two tasks.
\begin{table}[t]
\begin{center}
\begin{tabular}{|l|c|c|}
\hline
\bf Model & \bf IMDB (\%)  & \bf 20NewsGroup(\%) \\
\hline
Skip-gram & 87.06  & 77.20  \\
\hline
CBOW     & 87.20  & 73.22  \\
\hline
GloVe   & 85.68 & 68.06 \\
\hline
LDA     & 67.42 & 72.20 \\
\hline
TWE-1 & 83.50 & 76.43 \\
\hline
CeWE  & 87.69 & 77.56 \\
\hline
GCeWE  & \bf 88.56 & \bf 78.04 \\
\hline
\end{tabular}
\end{center}
\caption{\label{sim-table} Classification accuracy on IMDB and 20NewsGroup. The results of LDA for IMDB and 20NewsGroup are from (Maas et al., 2011) and (Liu et al., 2015 respectively.}
\end{table}
\begin{figure*}[t]
\centering
\includegraphics[width=6in,height=3.5in]{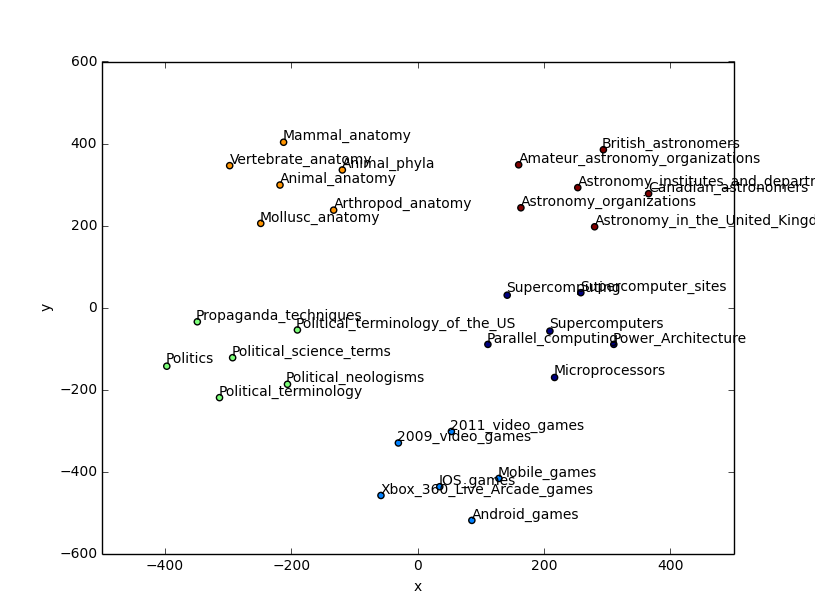}
\caption{Visualization of categories listed in the section of ``Qualitative Evaluation of Category Embeddings'' and their nearest categories.}
\end{figure*}

\subsection{\label{qece} Qualitative Evaluation of Category Embeddings}
To show that our learned category embeddings capture the topical
information, we randomly select 5 categories: supercomputers, IOS games, political terminology, animal anatomy, astronomy in the United Kingdom, and
compute the top 10 nearest words for each of them. For a given
category, we select words by comparing the cosine distance between
the category embedding and all other words in the vocabulary. Table
1 in the supplementary material lists words that have a distance to
the category embedding within the top 10 maximum distances.  For
example, given the category ``Animal Anatomy", it returns the
anatomical terminologies that are highly related to animal anatomy.
\indent We also project the embeddings of categories and words described above to the 2-dimensional space using the t-SNE
algorithm~\cite{tsne}, which is presented in Figure 1 in the
supplementary material. It is shown that
categories and corresponding neighbor words are projected into
similar positions, forming five clusters. Besides, we compute the 5
nearest categories for the categories listed above respectively and
we visualize it in Figure 3. As it can be seen, categories with
similar topical meanings
are projected into nearby positions.\\

\section{Conclusion and Future Work}
We have presented two models that integrate document category knowledge into
the learning of word embeddings and demonstrate the
ability of generalization of the learned word embeddings in several NLP tasks. For our future research work, we have plans to
integrate refined category knowledge and remove redundant categories
that may hinder the learning of word representations. We will
also consider how to leverage the learned category embeddings in other
NLP related tasks such as multi-label text classification.

\bibliography{catwv}
\bibliographystyle{naaclhlt2016}

\end{document}